\documentclass[runningheads]{llncs}

\usepackage[mobile]{eccv}

\usepackage{eccvabbrv}

\usepackage{booktabs}
\usepackage{amsmath}   
\usepackage{amssymb}  
\usepackage{mathtools} 
\usepackage[many]{tcolorbox}
\usepackage{xspace}   
 
\usepackage[table,xcdraw]{xcolor}
\definecolor{darkblue}{rgb}{0, 0, 0.5}
\definecolor{darkgreen}{RGB}{50,100,0}
\definecolor{darkred}{RGB}{200, 0, 0}
\definecolor{lightblue}{RGB}{220,235,250}
\usepackage{algorithm}
\usepackage{algpseudocode}

\definecolor{cvprblue}{rgb}{0.21,0.49,0.74}
\usepackage[pagebackref,breaklinks,colorlinks,allcolors=cvprblue]{hyperref}

\newcommand{\method}{SPINE\xspace}
\title{SPINE: Token-Selective Test-Time Reinforcement Learning with Entropy-Band Regularization}

\author{
Jianghao Wu$^{1,}$\thanks{Email: jianghao.wu@monash.edu},
Yasmeen George$^{1}$, 
Jin Ye$^{1}$, 
Yicheng Wu$^{2}$, \\
Daniel F. Schmidt$^{1}$, 
Jianfei Cai$^{1}$ \\
}

\institute{
Monash University \\
\and
Imperial College London \\
}

\begin{document}
\maketitle
\begin{abstract}
Large language models (LLMs) and multimodal LLMs (MLL-Ms) excel at chain-of-thought reasoning but face distribution shift at test-time and a lack of verifiable supervision. Recent test-time reinforcement learning (TTRL) methods derive
label-free pseudo-rewards from self-consistency voting over
sampled trajectories, yet they often collapse: the majority-vote reward prevails, responses shorten, and Pass@1 declines.
We trace this to uniform sequence updates in which most tokens are low-entropy followers, while a small high-entropy subset determines the reasoning branches.
Thus we propose \method, a token-selective test-time reinforcement learning framework that (i) performs distribution-aware forking-token selection to update only decision-critical branch points, and (ii) applies a robust entropy-band regularizer at those tokens to prevent premature collapse and suppress noisy drift. \method plugs into GRPO-style objectives (optionally with a KL anchor) and requires neither labels nor reward models.
Across eight benchmarks spanning multimodal VQA, text-only reasoning, \method consistently improves Pass@1 over TTRL while avoiding response-length collapse and yielding more stable training dynamics on both LLM and MLLM backbones. These results indicate that aligning updates with chain-of-thought branch points is a simple and label-free mechanism for stable and effective test-time adaptation in reasoning models. Code will be released.
\end{abstract}

\begin{figure*}[ht]
  \centering
  \includegraphics[width=\textwidth]{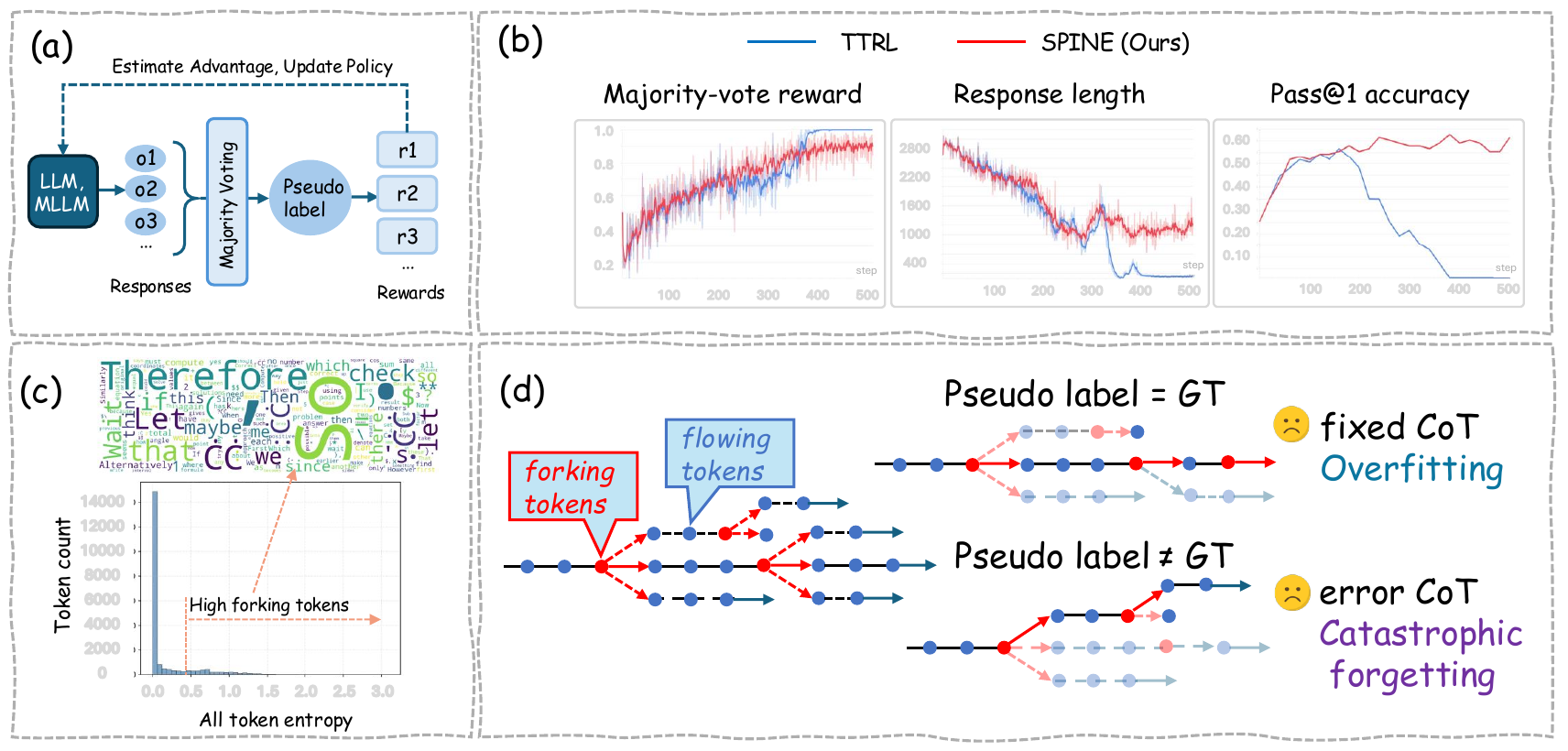}
\caption{Motivation of \method.
(a) TTRL: sample multiple responses, majority vote forms a pseudo-label, then update with GRPO.
(b) TTRL is unstable with shrinking outputs.
(c) Entropy is skewed; the high-entropy tokens mark forking decisions.
(d) \method updates only forking tokens and applies an entropy band, stabilizing adaptation and mitigating overfitting and forgetting.
}
  \label{fig:motivation_fig1}
\end{figure*}

\section{Introduction}\label{sec:intro}
Large-scale foundation models, including both language models (LLMs) and multimodal large language models (MLLMs), exhibit impressive chain-of-thought (CoT) reasoning across a wide range of general-domain tasks~\cite{wei2022chain,guo2025deepseek,jaech2024openai}. Yet real-world deployment faces two persistent pressures: \emph{distribution shift at test-time}~\cite{yuan2023revisiting,oh2025understanding,hu2025test} and \emph{the scarcity of verifiable supervision}~\cite{casper2023open,silver2025welcome}. Reinforcement learning with verifiable rewards (RLVR) can substantially improve reasoning~\cite{shao2024deepseekmath,lightman2023let}, but it presupposes dense labels or high-quality reward models that many domains lack, e.g., mathematical problem solving~\cite{liu2023improving}, clinical decision support~\cite{wang2023huatuo}, and scientific QA~\cite{hu2025survey}. These constraints motivate improving models directly on unlabeled test inputs rather than waiting for new annotations.

Test-Time Training (TTT) adapts models on incoming unlabeled data, typically via pseudo-labels or self-supervised signals~\cite{sun2019test,sun2024learning,behrouz2024titans,akyurek2024surprising}. However, recent evidence indicates that reinforcement learning generalizes more robustly than supervised fine-tuning (SFT) on reasoning tasks, where SFT often imitates surface patterns rather than improving deductive behavior~\cite{wu2025generalization,liu2025othink}. Building on this, Test-Time Reinforcement Learning (TTRL)~\cite{zuo2025ttrl} and unsupervised post-training for MLLMs~\cite{wei2025unsupervised} sample multiple reasoning paths and derive pseudo-rewards via self-consistency voting (Fig.~\ref{fig:motivation_fig1}a), yielding substantial gains without labels or reward models.
Nevertheless, in practice standard TTRL quickly develops a characteristic collapse mode. As updates proceed, the majority-vote reward keeps increasing while responses become shorter and Pass@1 eventually drops (Fig.~\ref{fig:motivation_fig1}b, top). This behavior suggests that the policy is optimizing agreement among sampled trajectories rather than correctness, converging to a small set of short, self-consistent but often incorrect answers. This collapse stems from learning on noisy pseudo-rewards: uniform sequence updates implicitly treat self-consistency as a faithful surrogate for correctness, thereby exposing a structural mismatch between the proxy signal and the true objective.

Recent analyses of token-entropy patterns in CoT under RLVR reveal a highly skewed distribution: most tokens are generated with low entropy, while only a small minority in the high-entropy tail exhibits substantial uncertainty (Fig.~\ref{fig:motivation_fig1}c). Prior work further shows that these high-entropy tokens often coincide with branch points that steer the downstream reasoning trajectory, motivating optimization on a fixed top proportion (e.g., 20\%) of high-entropy tokens~\cite{wang2025beyond80}. 
However, this perspective remains incomplete in the label-free TTRL setting. Under noisy self-consistency pseudo-rewards, the entropy profile can vary substantially across inputs, datasets, and adaptation steps. Consequently, the boundary between true \emph{forking tokens} and low-impact flowing tokens becomes inherently input-dependent and non-stationary, making a fixed top-$k\%$ rule brittle in practice. More importantly, identifying sparse decision-critical positions alone is not sufficient: even when updates are restricted to these tokens, their uncertainty may still collapse too early, pruning useful reasoning branches, or drift upward, amplifying pseudo-reward noise and destabilizing optimization. Taken together, these observations suggest that label-free TTRL must address two coupled challenges: \emph{where} to apply policy updates, and \emph{how} to maintain a stable uncertainty regime at those decision points.

We therefore propose \method (Selective Policy Improvements at Nodes of Entropy), a selective framework for test-time reinforcement learning. 
(i) {Distribution-Aware Forking Token Selection.} 
Instead of updating all tokens uniformly or relying on a fixed top-$k\%$ heuristic, \method adaptively identifies a small set of \emph{forking tokens} from the token-entropy distribution and applies GRPO-style policy updates only at these decision-critical positions, while preserving low-entropy flowing tokens to avoid perturbing low-uncertainty continuations.
(ii) {Robust Entropy-Band Regularization.} 
To further prevent collapse under noisy pseudo-rewards, \method explicitly regularizes the uncertainty of these forking tokens with a robust entropy band, increasing entropy when branching collapses too early and decreasing it when excessive uncertainty would amplify noisy supervision. 
\method reuses forward-pass statistics (log probabilities and token entropies), can incorporate a KL anchor, and requires neither labels nor external reward models.

Our main contributions can be summarized as follows.
\begin{itemize}
\item We identify a key limitation of label-free TTRL: under noisy self-consistency pseudo-rewards, both full-sequence updates and fixed-ratio high-entropy selection can yield misaligned or unstable policy improvement, since the set of decision-critical tokens is distribution-dependent and their uncertainty can still collapse or over-expand during adaptation.
\item We propose \method, a selective TTRL framework that combines distribution-aware \emph{forking token} selection with robust entropy-band regularization, enabling stable and targeted policy updates in the CoT decision space.
\item Across \emph{eight} benchmarks, \method consistently improves Pass@1 over standard TTRL on both LLM and MLLM backbones, while delivering more stable and reliable label-free test-time adaptation.
\end{itemize}

\begin{figure*}[ht]
  \centering
  \includegraphics[width=0.9\textwidth]{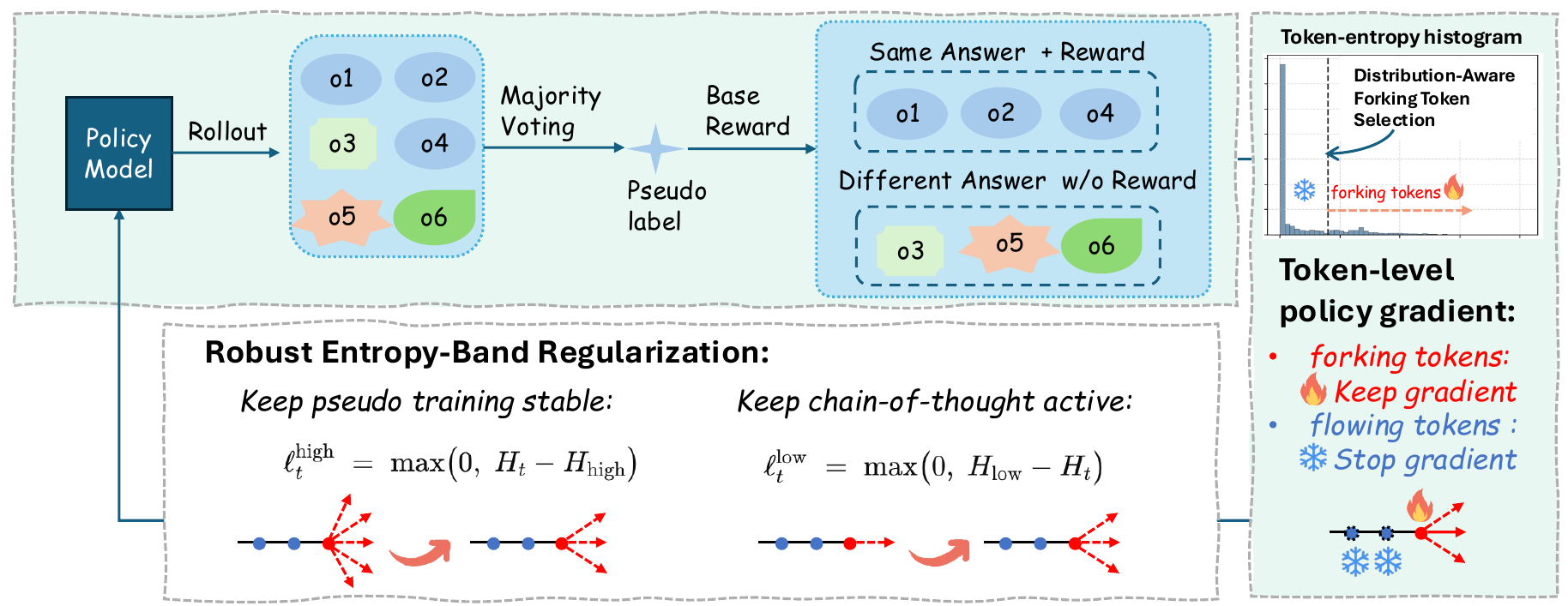}
\caption{\method pipeline. The model samples responses, majority voting produces a pseudo-label, and rewards are assigned. Gradients update only forking tokens, while flowing tokens are frozen. An entropy band further stabilizes training and preserves reasoning diversity.}
  \label{fig:pipleline_ours}
\end{figure*}

\section{Related Work}
\subsection{Reasoning in LLMs and MLLMs }
Reasoning in LLMs has been advanced mainly by supervised and self-supervised training that teach chain-of-thought (CoT), self-consistency, and reflection~\cite{wei2022chain,wang2022self}. Early LLMs internalize stepwise patterns via supervised fine-tuning (SFT) on (prompt, trace, answer) triplets, as in Flan-PaLM–style mixtures that enable zero-shot CoT~\cite{chung2024scaling,zelikman2022star}.
This SFT on CoT paradigm was then adopted by multimodal models. A common recipe first aligns vision and language through visual instruction tuning, for example, LLaVA and MiniGPT-4, and then performs SFT on multimodal chain-of-thought data to elicit stronger visual reasoning, for example, LLaVA CoT and related systems~\cite{liu2023visual,zhu2023minigpt,xu2025llava}. These methods report consistent gains across visual math and chart understanding benchmarks.
Despite its effectiveness, SFT behaves like behavior cloning of a single reasoning path, which can limit generalization and requires costly, high-quality reasoning traces~\cite{foster2024behavior,turpin2023language,chen2025towards}. 
These limitations motivate outcome-based reinforcement learning on tasks with verifiable solutions, where rewards are computed automatically by unit tests, checkers, or verifiers~\cite{le2022coderl,lightman2023let}. Within this line, GRPO is a practical objective for long CoT: it is critic-free, estimates advantages from groups of rollouts, and yields a stable low-variance signal~\cite{shao2024deepseekmath}. Building on GRPO, large reasoning models such as DeepSeekMath and the R1 series show strong gains on mathematical reasoning~\cite{shao2024deepseekmath,guo2025deepseek}, and related work extends outcome-based RL to multimodal settings to couple perception with stepwise reasoning~\cite{zhang2025r1,wei2025advancing}. Overall, reasoning benefits from SFT to bootstrap stepwise behavior and from outcome-based RL to move beyond imitation. However, many approaches still depend on reward models or human labels~\cite{ouyang2022training,lightman2023let}, which are costly in specialized domains and hard to maintain after deployment, motivating methods that learn from verifiable signals with minimal supervision at test time.

\subsection{Test-Time Scaling}
Test-time scaling increases the compute budget at inference without updating model parameters. Prior work suggests that for many reasoning tasks, allocating extra compute at test-time can be more sample efficient than scaling pretraining compute~\cite{snell2024scaling,liu2025can,kaplan2020scaling}. Two common forms are parallel generation and sequential generation~\cite{welleck2024decoding}.
Parallel generation draws multiple candidates or decision paths and then aggregates them, e.g., self-consistency and best of N~\cite{wang2022self,chen2023universal,stiennon2020learning,nakano2021webgpt}, Monte Carlo Tree Search for discrete decisions~\cite{zhou2023language,xie2024monte}, or token-level search, such as reward-guided sampling~\cite{deng2023reward,khanov2024args}. 
Aggregation may rely on simple voting or reward models~\cite{lightman2023let,wang2023math,zhangopenprm}. Sequential generation allocates more steps to a single response through reflection and chain-of-thought prompting~\cite{wei2022chain,madaan2023self}. While these strategies improve accuracy, their gains are ultimately bounded by the base model and by the cost and latency of large-scale sampling.
Beyond scaling inference-time sampling, {test-time training} (TTT) updates parameters on unlabeled inputs via pseudo-labels or self-supervision~\cite{akyurek2024surprising,wang2025test}. {Test-Time RL} (TTRL)~\cite{zuo2025ttrl} instead uses majority-vote self-consistency as a verifiable reward, with MM-UPT extending to multimodal models~\cite{wei2025unsupervised}. 
ETTRL~\cite{liu2025ettrl} reshapes rollouts and advantages via response-level entropy.
Compute as Teacher (CaT) remains label-free but introduces an external teacher/judge (e.g., GPT-4o) to synthesize and verify answers~\cite{jayalath2025compute}, thus replacing self-consistency with auxiliary model feedback. EVOL-RL is likewise label-free yet relies on an external embedding model to score novelty and performs full-sequence policy updates guided by embedding similarity~\cite{zhou2025evolving}.
In contrast, our method avoids external teachers and embedders and operates at the token level. We update only high-entropy \emph{forking tokens} and apply an entropy band with a masked KL at those positions, isolating the gains of selective token updates under a matched GRPO-style TTRL setup.

\section{Methodology}\label{sec:method}

\textbf{Setting and notation.}
We study test-time reinforcement learning for autoregressive reasoning models,
including both LLMs and MLLMs.
Given an input $x$, a parametric policy $\pi_\theta(y\mid x)$ generates an output
$y=(a_1,\dots,a_T)$ autoregressively, where each token is sampled as
$a_t \sim \pi_\theta(\cdot \mid s_t)$ and $s_t$ denotes the decoder state before emitting token $t$
(i.e., a summary of $x$ and $a_{<t}$).
During test-time adaptation, rollouts are sampled from a behavior policy $\pi_{\theta_{\mathrm{old}}}$, while the parameters $\theta$ are updated by GRPO
to obtain the new policy $\pi_\theta$.

At test time, the model receives unlabeled inputs and aims to improve its reasoning behavior without any ground-truth supervision.
We use self-consistency to derive label-free rewards, but standard GRPO with uniform token updates can be unstable under this noisy proxy.
\method addresses this issue by (i) updating only
distribution-aware \emph{forking tokens} and
(ii) regularizing their uncertainty with a robust
\emph{entropy band}, both within the GRPO objective.

\subsection{Self-consistency reward and GRPO objective}
\label{sec:self_consistency_grpo}

For each input $x$, we draw $N$ candidate responses $\{y_i\}_{i=1}^N \!\sim\! \pi_{\theta_{\mathrm{old}}}(\cdot\mid x)$
and aggregate them into a consensus output $y^\star$ (e.g., majority voting over extracted answers).
Each sampled response $y_i$ then receives a rule-based reward
\begin{equation}
r_i \;=\; r(y_i, y^\star)\in [0,1]
\end{equation}
This self-consistency reward encourages the model to prefer high-consensus outputs without relying on external supervision.

To optimize the policy under these rewards, we adopt
Grouped Relative Policy Optimization (GRPO), an on-policy
algorithm that replaces explicit value estimation with
group-wise normalized advantages. Within each group of $N$ samples for the same input $x$,
the standardized advantage for the $i$-th sample is computed as
\begin{equation}
\hat A^{\,i}
\;=\;
\frac{r_i - \operatorname{mean}(\{r_j\}_{j=1}^N)}
{\operatorname{std}(\{r_j\}_{j=1}^N) + \epsilon},
\qquad i = 1,\dots,N
\label{eq:grpo-adv}
\end{equation}
and the token-level PPO ratio is
\begin{equation}
\rho_t^{(i)}(\theta)
\;=\;
\frac{\pi_\theta\!\left(a_t^{(i)}\mid s_t^{(i)}\right)}
{\pi_{\theta_{\mathrm{old}}}\!\left(a_t^{(i)}\mid s_t^{(i)}\right)}.
\end{equation}
The clipped surrogate objective is then
\begin{equation}
\ell_{\mathrm{PPO},t}^{(i)}(\theta)\!=\!
\min\!\Big[
\rho_t^{(i)}\,\hat A^{\,i},\;
\operatorname{clip}(\rho_t^{(i)},\,1\!-\!\epsilon,\,1\!+\!\epsilon)\,\hat A^{\,i}
\Big]
\label{eq:ppo-clip}
\end{equation}
where the $\operatorname{clip}$ operator truncates the ratio to the interval $[1-\epsilon,\,1+\epsilon]$.

To ensure stable adaptation without over-regularizing non-forking positions, we apply a
token-level KL anchor only on forking tokens. Concretely, we define a masked,
size-normalized KL term
\begin{equation}
\ell_{\mathrm{KL}}^{\mathrm{fork}}
=
\frac{
\mathbb{E}_{(i,t)\in\mathcal B}\!\big[
m_t^{(i)}\,
D_{\mathrm{KL}}\!\big(
\pi_\theta(\cdot\mid s_t^{(i)})\,\Vert\,\pi_{\mathrm{ref}}(\cdot\mid s_t^{(i)})
\big)
\big]
}{
\mathbb{E}_{(i,t)\in\mathcal B}[\,m_t^{(i)}\,] + \epsilon
},
\label{eq:kl-masked}
\end{equation}
where $\pi_{\mathrm{ref}}$ denotes the fixed reference policy, set to the pre-adaptation base model $\pi_{\theta_0}$ and kept frozen throughout adaptation. 
Here
$m_t^{(i)}\!\in\!\{0,1\}$ masks the forking tokens selected by the distribution-aware thresholding in Sec.~\ref{sec:decision_tokens}. 

\subsection{Distribution-Aware Forking Token Selection}
\label{sec:decision_tokens}

Under noisy pseudo-rewards, uniform token updates dilute gradients away from decision-critical branch points.
Moreover, fixed top-$k\%$ selection is sensitive to entropy-scale shifts across prompts and update steps, which can over- or under-select tokens.
We therefore select \emph{forking tokens} via distribution-aware thresholding on token entropy. 
For each sampled response $y_i=(a^{(i)}_1,\dots,a^{(i)}_{T_i})$, we compute the token entropy of the current policy
\begin{equation}
H_t^{(i)}(\theta)
=
-\sum_{v\in\mathcal V}\pi_\theta(v\mid s_t^{(i)})\log \pi_\theta(v\mid s_t^{(i)}),
\qquad t=1,\dots,T_i .
\end{equation}
To decouple statistic estimation from optimization, we use a detached copy
\begin{equation}
\tilde H_t^{(i)}=\operatorname{sg}\!\left(H_t^{(i)}(\theta)\right),
\end{equation}
where $\operatorname{sg}(\cdot)$ denotes the stop-gradient operator.
This detached entropy is used only for token selection and band construction, while the original $H_t^{(i)}(\theta)$ is retained for entropy regularization.

We build a histogram of $\{\tilde H_t^{(i)}\}_{t=1}^{T_i}$ with $B$ bins (fixed to $B{=}100$).
Let $p_j^{(i)}$ be the normalized mass of bin $j$ (so $\sum_{j=1}^{B} p_j^{(i)}=1$) and $c_j^{(i)}$ its center.
For a candidate split $k\in\{1,\dots,B\!-\!1\}$, define
\begin{equation}
\omega_0^{(i)}(k)=\sum_{j=1}^{k} p_j^{(i)},\quad
\omega_1^{(i)}(k)=\sum_{j=k+1}^{B} p_j^{(i)},
\end{equation}
\begin{equation}
\mu_0^{(i)}(k)=\frac{1}{\omega_0^{(i)}(k)}\sum_{j=1}^{k} p_j^{(i)}c_j^{(i)},\quad
\mu_1^{(i)}(k)=\frac{1}{\omega_1^{(i)}(k)}\sum_{j=k+1}^{B} p_j^{(i)}c_j^{(i)}.
\end{equation}
We consider only valid splits with $\omega_0^{(i)}(k)>0$ and $\omega_1^{(i)}(k)>0$.
Otsu's criterion selects the split that maximizes the between-class variance:
\begin{equation}
k^\star=\arg\max_{k}\;\omega_0^{(i)}(k)\,\omega_1^{(i)}(k)\big(\mu_0^{(i)}(k)-\mu_1^{(i)}(k)\big)^2,
\qquad
\tau^{(i)} = c_{k^\star}^{(i)} .
\end{equation}
If no valid split exists (e.g., extremely short or near-degenerate entropy distributions), we set
\begin{equation}
\tau^{(i)}=\max_{t\in\{1,\dots,T_i\}} \tilde H_t^{(i)}.
\end{equation}
We then define the forking-token set and mask by thresholding:
\begin{equation}
\mathcal S_i=\{t\in\{1,\dots,T_i\}\mid \tilde H_t^{(i)}\ge \tau^{(i)}\},\qquad
m_t^{(i)}=\mathbf{1}\!\left[\tilde H_t^{(i)}\ge \tau^{(i)}\right].
\end{equation}

\subsection{Robust Entropy-Band Regularization}
\label{sec:entropy_band}

Selecting forking tokens specifies \emph{where} to update.
However, under noisy self-consistency rewards, the uncertainty at these tokens can either collapse prematurely
(leading to branch pruning and reward saturation) or drift upward
(amplifying sampling noise and destabilizing advantage estimation).
We therefore regulate token-level uncertainty at forking positions
by constraining the entropy of the current policy to remain within a robust, data-driven band.

For each sample $i$, we collect the detached entropies of forking tokens:
\begin{equation}
\tilde{\mathcal E}_i
=
\{ \tilde H_t^{(i)} \mid t \in \mathcal S_i \}.
\end{equation}

We estimate the central tendency and scale of $\tilde{\mathcal E}_i$
using the median and median absolute deviation (MAD):
\begin{equation}
\mu_i=\operatorname{median}(\tilde{\mathcal E}_i),\qquad
\mathrm{MAD}_i=\operatorname{median}\big(\{\,|\tilde H_t^{(i)}-\mu_i| \mid t\in\mathcal S_i\,\}\big).
\end{equation}

A robust scale estimate is obtained as
\begin{equation}
s_i
=
\max\big(1.4826 \cdot \mathrm{MAD}_i,\; 10^{-6}\big),
\end{equation}
where $1.4826$ ensures consistency with standard deviation under Gaussian assumptions.
We define an asymmetric entropy band as
\begin{equation}
H_{\mathrm{high}}^{(i)} = \operatorname{sg}(\mu_i),
\qquad
H_{\mathrm{low}}^{(i)} = \operatorname{sg}\!\big(\max(0,\; \mu_i - s_i)\big).
\end{equation}
The upper bound is set to the median, while the lower bound
is relaxed by one robust scale. This asymmetric design reflects
our emphasis on stability in unsupervised adaptation:
upward entropy drift is penalized more strictly to avoid
amplifying noisy pseudo-rewards.
Violations are penalized via hinge losses on the original entropy:
\begin{align}
\ell_t^{\mathrm{high}}
&=
\max\!\big(
0,\,
H_t^{(i)}(\theta) - H_{\mathrm{high}}^{(i)}
\big), \\
\ell_t^{\mathrm{low}}
&=
\max\!\big(
0,\,
H_{\mathrm{low}}^{(i)} - H_t^{(i)}(\theta)
\big).
\end{align}

The overall entropy-band regularizer is
\begin{equation}
\mathcal R_{\mathrm{band}}(\theta)
=
\mathbb{E}_{(i,t)\in\mathcal B}
\Big[
\big(
\beta_\ell \ell_t^{\mathrm{low}}
+
\beta_u \ell_t^{\mathrm{high}}
\big)
\, m_t^{(i)}
\Big].
\end{equation}

\subsection{Final Objective}
\label{sec:final_obj}

Combining token-selective updates and entropy-band regulation,
we optimize the following objective over mini-batches $\mathcal B$:
\begin{equation}
\mathcal{L}_{\mathrm{core}}
=
-
\mathbb{E}_{(i,t)\in\mathcal{B}}
\big[
m_t^{(i)}\,\ell_{\mathrm{PPO},t}^{(i)}(\theta)
\big]
+
\lambda_{\mathrm{KL}}
\,
\ell_{\mathrm{KL}}^{\mathrm{fork}},
\end{equation}
where $\ell_{\mathrm{PPO},t}^{(i)}(\theta)$ is defined in Eq.~\eqref{eq:ppo-clip}
and $\ell_{\mathrm{KL}}^{\mathrm{fork}}$ in Eq.~\eqref{eq:kl-masked}.
The mask $m_t^{(i)}$ ensures that policy updates are applied only at forking tokens.

The final loss incorporates the entropy-band regularizer:
\begin{equation}
\mathcal{L}
=
\mathcal{L}_{\mathrm{core}}
+
\mathcal{R}_{\mathrm{band}}.
\end{equation}
Together, these components yield stable test-time reinforcement learning under label-free and noisy self-supervision.

\begin{table*}[!t]
\centering
\caption{
Performance of \method on multimodal reasoning benchmarks.
\method consistently improves over TTRL and the base model across multimodal VQA tasks (MathVision, SLAKE, MedXpertQA-MM).
}
\label{tab:results_vqa_medqa}
\resizebox{0.80\linewidth}{!}{
\begin{tabular}{l|ccc|c}
\toprule
\textbf{Name} & \textbf{MathVision} & \textbf{SLAKE} & \textbf{MedXpertQA-MM}  & \textbf{Avg} \\
\midrule
\textbf{Base Model} & \multicolumn{3}{c}{\textbf{Qwen2.5-VL-3B-Instruct}}  & \\
\midrule
No adaptation & $19.65$ & $26.17$ & $17.17$ & $21.00$ \\
Self-Consistency & 19.20 & 25.84 & 19.47 & 21.50 \\
w/  LMSI & 9.21 & 9.05 & 22.01 & 13.42 \\
w/   SEALONG & 10.36 & 12.32 & 6.51 & 9.73 \\
\rowcolor{lightblue!100} w/ TTRL & $22.73$ & $30.00$ & $22.61$ & $25.11$ \\
\rowcolor{lightblue!100} w/ \method{} & $27.18$ & $32.62$ & $23.84$ & $27.88$ \\
\rowcolor{lightblue!100} $\Delta$ & \textcolor{red}{$+4.5$} & \textcolor{red}{$+2.6$} & \textcolor{red}{$+1.2$} & \textcolor{red}{$+2.8$} \\
\bottomrule
\end{tabular}}
\vspace{-4mm}
\end{table*}

\begin{table*}[!t]
\centering
\caption{
Performance of \method on mathematical, general, and expert reasoning benchmarks. 
\method consistently outperforms both TTRL and the base model across mathematical tasks (AIME 2025, AMC, MATH-500) and general or expert benchmarks (GPQA, MMLU).
}
\label{tab:result_math_genral}
\begin{tabular}{l|ccccc|c}
\toprule
\textbf{Name} & \textbf{AIME 2025} & \textbf{AMC} & \textbf{MATH-500} & \textbf{GPQA} & \textbf{MMLU} & \textbf{Avg} \\
\midrule
\textbf{Base Model} & \multicolumn{5}{c}{\textbf{Qwen2.5-Math-1.5B}} \\
\midrule
No adaptation & $10.00$ & $28.91$ & $30.20$ & $4.06$ & - & $18.29$ \\
 Self-Consistency & 3.33 & 31.02 & 45.37 & 6.15 & - & 21.47 \\
w/  LMSI & 6.67 & 26.50 & 31.50 & 19.19 & - & 20.97 \\
 w/  SEALONG & 6.67 & 25.30 & 32.60 & 18.69 & - & 20.82 \\
\rowcolor{lightblue!100} w/ TTRL & $16.67$ & $49.88$ & $66.42$ & $25.38$ & - & $39.59$ \\
\rowcolor{lightblue!100} w/ \method{} & $23.33$ & $54.22$ & $74.00$ & $28.93$ & - & $45.12$ \\
\rowcolor{lightblue!100} $\Delta$ & \textcolor{red}{$+6.7$} & \textcolor{red}{$+4.3$} & \textcolor{red}{$+7.6$} & \textcolor{red}{$+3.6$} & \textcolor{red}{-} & \textcolor{red}{$+5.5$} \\

\midrule
\textbf{Base Model} & \multicolumn{5}{c}{\textbf{Qwen3-1.7B}} \\
\midrule
No adaptation & $10.00$ & $25.30$ & $71.60$ & $9.09$ & $58.16$ & $34.83$ \\
Self-Consistency & 10.83 & 33.13 & 66.07 & 8.75 & 65.39 & 36.83 \\
w/  LMSI & 16.67 & 34.94 & 62.40 & 18.18 & 71.74 & 40.78 \\
 w/  SEALONG & 20.00 & 40.96 & 61.00 & 13.64 & 69.36 & 40.99 \\
\rowcolor{lightblue!100} w/ TTRL & $26.67$ & $56.63$ & $79.86$ & $29.94$ & $71.19$ & $52.86$ \\
\rowcolor{lightblue!100} w/ \method{} & $36.67$ & $62.65$ & $82.80$ & $35.55$ & $72.48$ & $58.03$ \\
\rowcolor{lightblue!100} $\Delta$ & \textcolor{red}{$+10.0$} & \textcolor{red}{$+6.0$} & \textcolor{red}{$+2.9$} & \textcolor{red}{$+5.6$} & \textcolor{red}{$+1.3$} & \textcolor{red}{$+5.2$} \\
\bottomrule
\end{tabular}
\vspace{-4mm}
\end{table*}

\begin{table}[!t]
\centering
\caption{
Cross-Task Generalization of \method based on Qwen3-1.7B. 
Each model is adapted on one dataset and evaluated across all four benchmarks to measure generalization and forgetting.
}
\label{tab:generalization_and_forgetting}
\resizebox{0.8\linewidth}{!}{
\begin{tabular}{l|cccc|c}
\toprule
\textbf{Training} & \textbf{AIME 2025} & \textbf{AMC} & \textbf{MATH-500} & \textbf{GPQA} & \textbf{Avg} \\
\midrule
Qwen3-1.7B & $10.00$ & $25.30$ & $71.60$ & $9.09$ & $29.00$ \\
\hline
AIME 2025 & $36.67$ & $56.63$ & $80.60$ & $23.86$ & $49.44$ \\
\rowcolor{lightblue!100} $\Delta$ &
\textcolor{red}{$+26.7$} & \textcolor{red}{$+31.3$} & \textcolor{red}{$+9.0$} & \textcolor{red}{$+14.8$} & \textcolor{red}{$+20.4$} \\

AMC & $10.00$ & $62.65$ & $76.60$ & $29.95$ & $44.80$ \\
\rowcolor{lightblue!100} $\Delta$ &
\textcolor{red}{$+0.0$} & \textcolor{red}{$+37.4$} & \textcolor{red}{$+5.0$} & \textcolor{red}{$+20.9$} & \textcolor{red}{$+15.8$} \\

MATH-500 & $20.00$ & $51.81$ & $82.80$ & $21.83$ & $44.11$ \\
\rowcolor{lightblue!100} $\Delta$ &
\textcolor{red}{$+10.0$} & \textcolor{red}{$+26.5$} & \textcolor{red}{$+11.2$} & \textcolor{red}{$+12.7$} & \textcolor{red}{$+15.1$} \\

GPQA & $16.67$ & $54.22$ & $79.20$ & $35.55$ & $46.41$ \\
\rowcolor{lightblue!100} $\Delta$ &
\textcolor{red}{$+6.7$} & \textcolor{red}{$+28.9$} & \textcolor{red}{$+7.6$} & \textcolor{red}{$+26.5$} & \textcolor{red}{$+17.4$} \\
\bottomrule
\end{tabular}}
\vspace{-4mm}
\end{table}

\section{Experiments}\label{sec:experiments}

\subsection{Experimental Setup}
\textbf{Models}
We evaluate \method on a compact, representative suite spanning multimodal and text-only LLMs, covering model {type} (MLLM vs.\ LLM), {specialization} (generalist vs.\ math-focused), and {scale} (1.5–3B). Concretely, we use Qwen2.5-VL-3B-Instruct for multimodal reasoning~\cite{qwen2.5}, Qwen3-1.7B as a general-purpose text-only LLM~\cite{yang2025qwen3}, and Qwen2.5-Math-1.5B as a math-specialized LLM~\cite{yang2024qwen2-math}. 
Experiments initialize from publicly released checkpoints.

\textbf{Benchmarks}
We evaluate \method across three task families. For multimodal VQA, we consider {MathVision}~\cite{wang2024mathvision} for diagram-based mathematical reasoning, as well as {SLAKE}~\cite{liu2021slake} and {MedXpertQA-MM}~\cite{zuo2025medxpertqa} for clinical image understanding. For general and expert knowledge QA, we include {GPQA}~\cite{rein2024gpqa} and {MMLU}~\cite{hendrycks2020mmlu}. For mathematical reasoning, we use {AIME 2025}, {AMC}, and the {MATH-500} subset of MATH~\cite{hendrycks2021measuring}.

\textbf{Baselines}
We use standard TTRL (GRPO with a KL anchor and majority-vote self-consistency)~\cite{zuo2025ttrl} as our primary baseline, and additionally report No adaptation and Self-Consistency (majority vote without updates). For broader comparison, we include two self-improvement baselines: LMSI~\cite{huang2023lmsi}, which generates high-confidence CoT pseudo-labels via self-consistency and performs supervised fine-tuning; and SEALONG~\cite{li2024sealong}, which samples multiple long-context trajectories, scores them via MBR-style consensus, and fine-tunes on the top outputs.

\textbf{Implementation Details and Evaluation Protocol}
We implement \method\ with GRPO on all benchmarks. During adaptation, for each prompt we sample \(N{=}8\) rollouts with temperature \(0.7\) and top-\(p{=}0.95\), aggregate a pseudo-label via self-consistency (majority vote), and update the policy using GRPO with an optional KL anchor to the base model. At evaluation, we use greedy decoding (temperature \(0\)) and report {Pass@1} accuracy (a single greedy output). Predictions are matched to references after a standard normalization pipeline, case folding, whitespace cleanup, Unicode/LaTeX canonicalization (including symbol mapping), unit-word removal, mixed-number handling, and algebraic equivalence checking via \texttt{sympy} when applicable; full rules follow  \texttt{grade\_answer} implementation. We set the maximum output length to \(3072\) tokens for LLM tasks and \(2048\) tokens for multimodal tasks, and keep sampling filters and stopping criteria identical across methods. For multimodal models, following MM-UPT~\cite{wei2025unsupervised}, we do not freeze the vision tower during training. All runs use a fixed seed for reproducibility and are conducted on \(4\times\)NVIDIA A100\,80GB GPUs. For fair comparison, all experiments are conducted using the EasyR1 framework~\cite{zheng2025easyr1}
 under a unified configuration.
Except for the hyperparameters specific to our method, all other training settings are kept identical across experiments.

\subsection{Main Results}\label{sec:main_results}

\textbf{Results on multimodal VQA tasks.}
Table~\ref{tab:results_vqa_medqa} shows that TTRL improves the no-adaptation baseline on MathVision, SLAKE, and MedXpertQA-MM from 19.65/26.17/17.17 to 22.73/30.00/22.61. Building on this, \method further increases performance to 27.18/32.62/23.84, yielding gains of +4.5, +2.6, and +1.2 over TTRL, respectively (Avg: 27.88 vs.\ 25.11, +2.8).
In contrast, recent SFT-based methods such as LMSI and SEALONG do not exhibit consistent improvements and even lead to marked drops on MathVision and SLAKE, highlighting the limited generalization of supervised fine-tuning under unseen multimodal distributions.

\textbf{Results on mathematical and general/expert reasoning tasks.}
Table~\ref{tab:result_math_genral} summarizes the performance on three mathematical benchmarks (AIME 2025, AMC, MATH-500) and two general/expert benchmarks (GPQA, MMLU).
On Qwen2.5-Math-1.5B, TTRL already brings substantial improvements over the no-adaptation baseline, boosting AIME 2025 from 10.00 to 16.67, AMC from 28.91 to 49.88, and MATH-500 from 30.20 to 66.42; it also improves GPQA from 4.06 to 25.38.
Building on this strong TTRL baseline, \method delivers consistent additional gains across all reported tasks for this model, reaching 23.33/54.22/74.00 on AIME 2025/AMC/MATH-500 and 28.93 on GPQA.
This corresponds to further improvements over TTRL of +6.7, +4.3, +7.6, and +3.6, respectively, and increases the overall average from 39.59 to 45.12 (+5.5).

The same trend is observed on Qwen3-1.7B.
Compared with TTRL, \method improves AIME 2025 from 26.67 to 36.67 (+10.0), AMC from 56.63 to 62.65 (+6.0), and MATH-500 from 79.86 to 82.80 (+2.9).
Importantly, the benefits extend beyond math: \method also raises GPQA from 29.94 to 35.55 (+5.6) and MMLU from 71.19 to 72.48 (+1.3), yielding an average gain of +5.2 (58.03 vs.\ 52.86).
In contrast, SFT-based approaches such as LMSI and SEALONG offer only mild improvements over the base model and remain clearly behind TTRL and \method.

Overall, these results indicate that selective token updates with entropy-band regularization provide a more effective and reliable adaptation mechanism than TTRL, improving both mathematical reasoning and general/expert performance.

\begin{table}[!t]
\centering
\caption{
Ablation study of \method{} on three representative benchmarks.
\textbf{FT} denotes forking-token selective updates (either fixed-ratio or Otsu-adaptive),
and \textbf{EB} denotes entropy-band regularization.
}
\label{tab:ablationstudy}
\resizebox{0.80\linewidth}{!}{
\begin{tabular}{l|ccc|c}
\toprule
\textbf{Method} & \textbf{MathVision} & \textbf{AIME25} & \textbf{AMC} & \textbf{Avg} \\
\midrule
Base & 19.65 & 10.00 & 25.30 & 18.32 \\
\hline
TTRL & 22.73 & 26.67 & 56.63 & 35.34 \\
\rowcolor{lightblue!100} $\Delta$ vs.\ Base
& \textcolor{red}{+3.1} & \textcolor{red}{+16.7} & \textcolor{red}{+31.3} & \textcolor{red}{+17.0} \\
\hline
TTRL + FT (Top-20\%) & 25.12 & 26.67 & 59.03 & 36.94 \\
\rowcolor{lightblue!100} $\Delta$ vs.\ Base
& \textcolor{red}{+5.5} & \textcolor{red}{+16.7} & \textcolor{red}{+33.7} & \textcolor{red}{+18.6} \\
TTRL + FT (Otsu) & 25.49 & 30.00 & 60.24 & 38.58 \\
\rowcolor{lightblue!100} $\Delta$ vs.\ Base
& \textcolor{red}{+5.8} & \textcolor{red}{+20.0} & \textcolor{red}{+34.9} & \textcolor{red}{+20.3} \\
\hline
\method{} (FT+EB) & 27.18 & 36.67 & 62.65 & 42.17 \\
\rowcolor{lightblue!100} $\Delta$ vs.\ Base
& \textcolor{red}{+7.5} & \textcolor{red}{+26.7} & \textcolor{red}{+37.4} & \textcolor{red}{+23.9} \\
\bottomrule
\end{tabular}}
\vspace{-4mm}
\end{table}

\textbf{\method generalizes well beyond the target task.}
To assess whether \method overfits the adaptation dataset or suffers from cross-task forgetting, we conduct a cross-task evaluation on four benchmarks using Qwen3-1.7B. As shown in Table~\ref{tab:generalization_and_forgetting}, adapting on a single dataset consistently improves performance on the other unseen benchmarks, leading to large average gains over the no-adaptation baseline (Avg: 29.00). 
In particular, adapting on AIME 2025 achieves the strongest overall improvement, raising the average Pass@1 from 29.00 to 49.44 (+20.4) while simultaneously improving AMC, MATH-500, and GPQA by +31.3, +9.0, and +14.8, respectively.
Adapting on AMC and MATH-500 exhibits a similar trend: despite being optimized on one target benchmark, the resulting models still transfer well, improving the average to 44.80 (+15.8) and 44.11 (+15.1), with consistent gains on the remaining tasks.
Finally, adapting on GPQA yields the largest in-domain gain on GPQA (+26.5) and maintains positive transfer to all three math benchmarks (+6.7 on AIME 2025, +28.9 on AMC, and +7.6 on MATH-500), improving the average to 46.41 (+17.4).
Overall, \method demonstrates strong cross-task generalization under label-free adaptation, with no evident catastrophic forgetting on unseen benchmarks.

\subsection{Ablation Study}
\textbf{Component Analysis.}
Table~\ref{tab:ablationstudy} reports the ablation results of \method{} on MathVision, AIME25, and AMC.
TTRL already provides a strong boost over the base model, improving the average from 18.32 to 35.34 (+17.0).
Adding forking-token selective updates (FT) yields further gains.
A fixed top-20\% selection improves the average to 36.94, while the proposed Otsu-adaptive FT is consistently stronger, reaching 38.58 (+20.3 vs.\ base).
Finally, combining adaptive FT with entropy-band regularization (i.e., \method{}) achieves the best overall performance, raising the average to 42.17 (+23.9 vs.\ base and +6.8 vs.\ TTRL).
These results confirm that distribution-aware token selection is critical for effective adaptation, and that entropy-band regularization further improves robustness beyond selection alone.

\section{Analysis and Discussions}\label{sec:discussions}
\begin{figure}
    \centering
    \includegraphics[width=1.\linewidth]{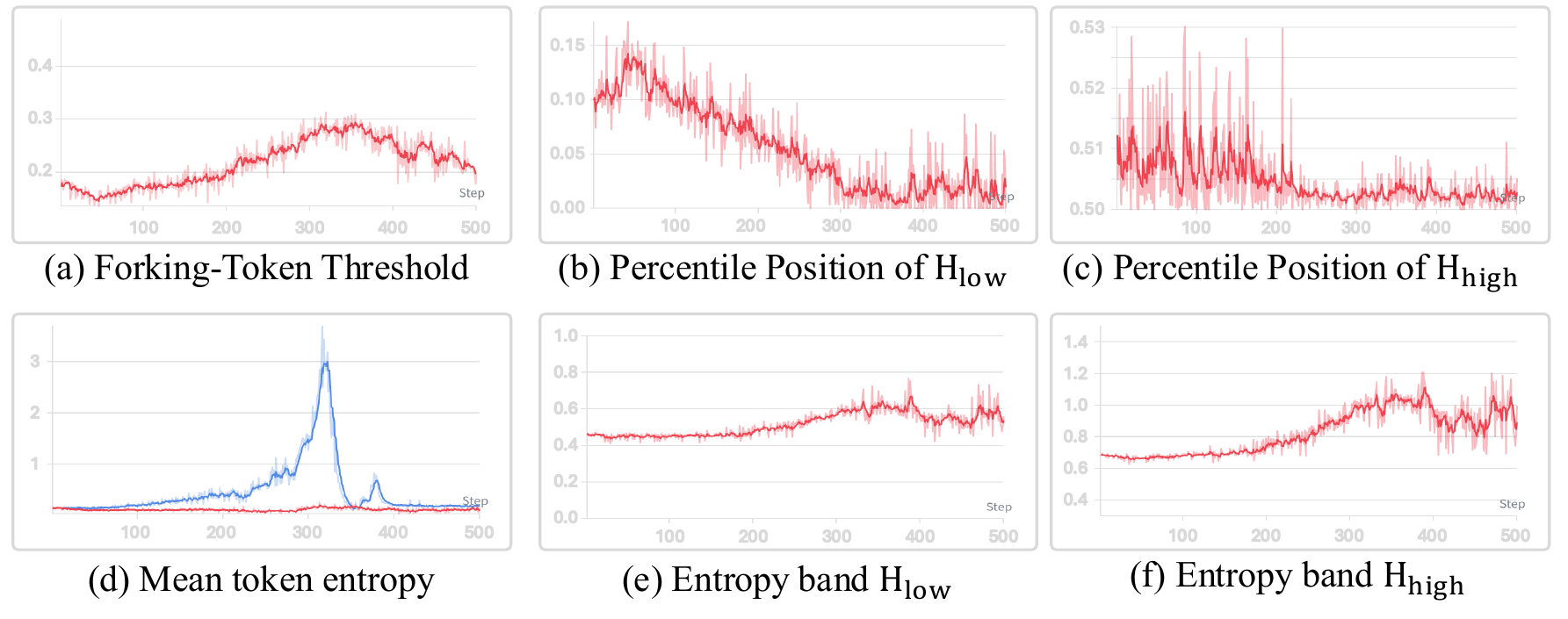}
\caption{\textbf{Training dynamics of SPINE on AMC.}
(a) Adaptive forking-token threshold $\tau$ from distribution-aware entropy splitting.
(b--c) Percentile positions of $H_{\mathrm{low}}$ and $H_{\mathrm{high}}$ in the token-entropy distribution.
(d) Mean token entropy over training.
(e--f) Absolute entropy-band bounds used by SPINE.
\textcolor{blue}{Blue}: \method{}. \textcolor{red}{Red}: TTRL.}
    \label{fig:dynamic_training}
        \vspace{-4mm}
\end{figure}

\textbf{Training Dynamics.}
We revisit the collapse of test-time reinforcement learning in Fig.~\ref{fig:motivation_fig1}(b).
Under TTRL, the majority-vote reward quickly saturates and responses shorten, followed by a drop in Pass@1, suggesting overfitting to pseudo-consensus.
Fig.~\ref{fig:dynamic_training}(d) reveals a matching entropy signature: mean token entropy rises, spikes sharply, and then collapses, indicating unstable uncertainty under noisy pseudo-rewards.
In contrast, \method maintains a controlled entropy regime throughout training.
As shown in Fig.~\ref{fig:dynamic_training}(a--c), the forking-token threshold $\tau$ and the percentile positions of $H_{\mathrm{low}}$/$H_{\mathrm{high}}$ adapt to the evolving entropy distribution, while the absolute band bounds remain well-behaved (Fig.~\ref{fig:dynamic_training}(e--f)).
This adaptive thresholding and band control prevent both entropy drift and premature collapse, leading to more stable trajectories and consistently higher Pass@1 than TTRL (Fig.~\ref{fig:motivation_fig1}(b)). \\

\begin{figure}
    \centering
    \includegraphics[width=0.9\linewidth]{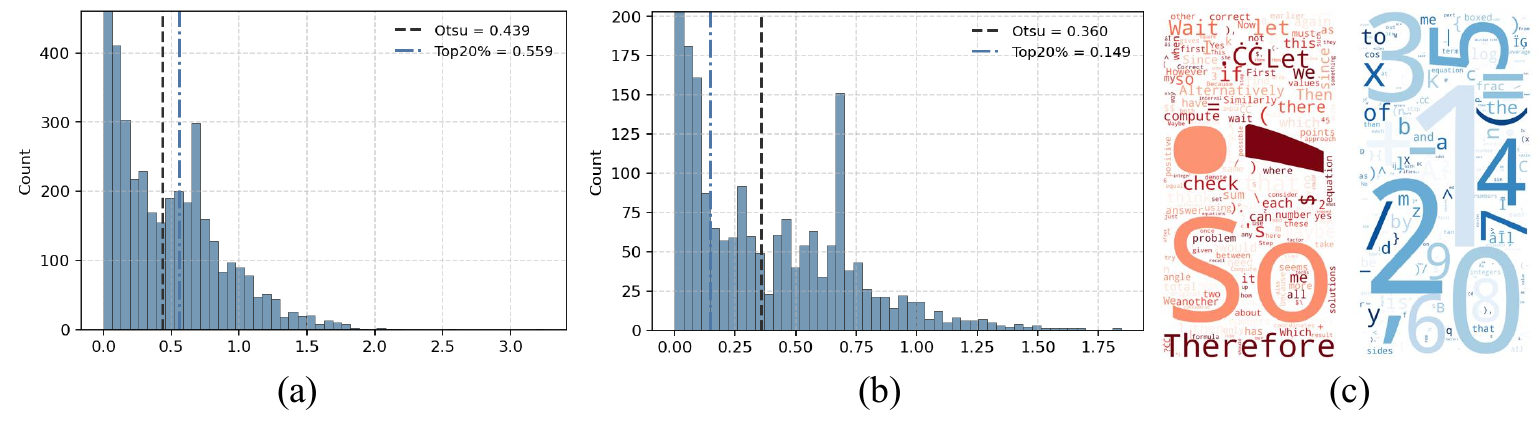}
\caption{
(a--b) Token-entropy distributions with Otsu vs.\ top-20\% thresholds.
(c) Word clouds: selected high-entropy tokens (left) and remaining tokens (right).}
    \label{fig:word-cloud}
        \vspace{-4mm}
\end{figure}

\textbf{Token-entropy distributions.} Figure~\ref{fig:word-cloud} shows that token entropy is heavy-tailed but varies in scale across samples, making a fixed Top-20\% rule unstable. Otsu provides a sample-adaptive split that better isolates the high-entropy tail, and the selected tokens concentrate on decision/branching cues (Fig.~\ref{fig:word-cloud}c), consistent with forking-token updates.

\begin{figure}
    \centering
    \includegraphics[width=1.\linewidth]{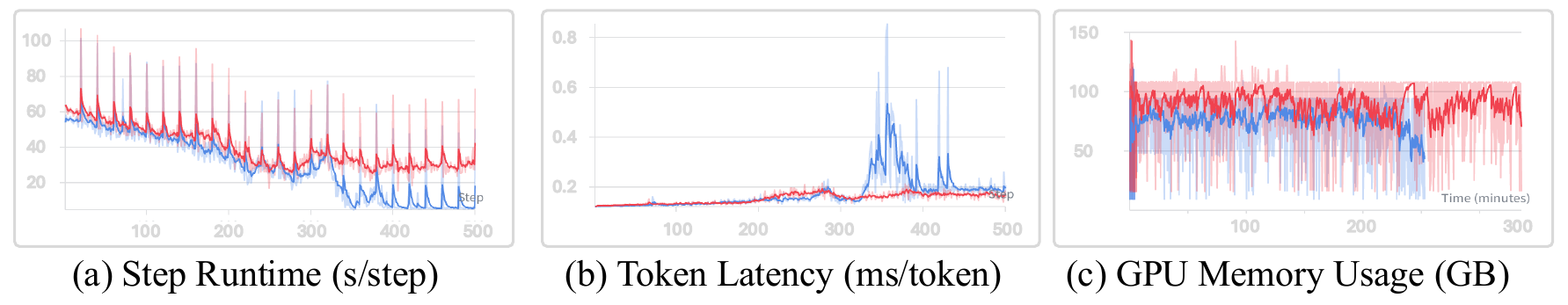}
\caption{\textbf{Efficiency comparison between SPINE and TTRL on GPQA.}
(a) Runtime per optimization step (s/step).
(b) Token-level generation latency (ms/token).
(c) GPU memory usage during adaptation (GB).
\textcolor{blue}{Blue}: \method{}. \textcolor{red}{Red}: TTRL.}
    \label{fig:computering}
        \vspace{-4mm}
\end{figure}

\textbf{Computational Cost.}
\label{sec:efficiency}
We analyze the computational overhead of \method in Fig.~\ref{fig:computering}. 
\method incurs a slightly higher step runtime than TTRL early on, mainly due to entropy-band computation and token-selective updates. 
The runtime gap widens later because TTRL collapses to shorter responses (Fig.~\ref{fig:motivation_fig1}(b)), reducing per-step work, whereas \method maintains stable response lengths and thus a steadier runtime profile. 
Token latency is comparable for most of training, but increases for TTRL after collapse due to unstable generation, while \method remains stable. 
\method uses slightly more GPU memory due to storing entropy statistics and a larger KV cache from longer responses, but the overhead remains modest in practice.

\textbf{Limitations and Failure Modes.}\label{sec:limitation_fail}
\method relies on self-consistency voting to construct pseudo-rewards. When the model is systematically biased under severe domain mismatch, the consensus may be consistently incorrect, yielding saturated yet uninformative rewards; in this regime, entropy control can stabilize training but cannot provide a corrective learning signal, and adaptation may reinforce spurious solutions.
\method also assumes that high-entropy tokens correlate with decision-critical branch points. For tasks where entropy is poorly calibrated, the selected forking tokens may be less aligned with true causal decisions, reducing the benefit of token-selective updates.
Finally, although \method reuses forward-pass statistics and updates only a subset of tokens, it introduces additional computation and slightly higher memory usage, which may be undesirable in strict latency- or memory-constrained deployments.

\section{Conclusion}
We presented \method, a token-selective test-time reinforcement learning framework for label-free adaptation of autoregressive reasoning models.
SPINE addresses a key failure mode of standard TTRL: uniform sequence updates overfit to noisy self-consistency rewards, leading to majority-vote saturation, response-length collapse, and degraded Pass@1.
Our approach combines (i) distri-bution-aware forking-token selection to focus updates on decision-critical branch points and (ii) a robust entropy-band regularizer to prevent entropy collapse and excessive drift.
Across eight benchmarks spanning multimodal VQA, text-only reasoning , \method consistently outperforms TTRL and no-adaptation baselines in practice while robustly avoiding majority-vote saturation, response-length collapse, and entropy drift. Empirically, analyses of training dynamics and token-entropy distributions indicate that \method preserves diverse reasoning trajectories and concentrates learning where it most affects the answer.

\newpage

\bibliographystyle{splncs04}
\bibliography{main}

\end{document}